# LLM-Measure: Generating Valid, Consistent, and Reproducible Text-Based Measures for Social Science Research

Yi Yang[1]*, Hanyu Duan[1], Jiaxin Liu[1], Kar Yan Tam[1]

[1]Department of Information Systems, Business Statistics and Operations Management, Hong Kong University of Science and Technology; Sai Kung, Hong Kong SAR.

*Corresponding author. Email: imyiyang@ust.hk

**Abstract:** The increasing use of text as data in social science research necessitates the development of valid, consistent, reproducible, and efficient methods for generating text-based concept measures. This paper presents a novel method that leverages the internal hidden states of large language models (LLMs) to generate these concept measures. Specifically, the proposed method learns a concept vector that captures how the LLM internally represents the target concept, then estimates the concept value for text data by projecting the text's LLM hidden states onto the concept vector. Three replication studies demonstrate the method's effectiveness in producing highly valid, consistent, and reproducible text-based measures across various social science research contexts, highlighting its potential as a valuable tool for the research community.

**Main Text:**

Text is data. The rapid digitalization and increasing prevalence of text data have sparked growing interest in using it to address important societal and economic questions in social science research (*1–3*). For example, in macroeconomics, how can the Federal Reserve's (the Fed) monetary policy stance be gauged from the Federal Open Market Committee (FOMC) minutes? In microeconomics, how can a company's innovation culture be measured from its corporate disclosures? And in consumer behavior, how can the information overload in online consumer reviews be assessed?

Answering these questions often requires constructing a concept derived from text data, which can then be used in quantitative analysis. Researchers have traditionally relied on natural language processing (NLP) techniques, such as keyword and rule-based methods, as well as machine learning approaches. While keyword and rule-based methods are simple, they require significant expert effort to create and adapt, and are slow to adjust to new contexts. Machine learning models, though more advanced, often rely on human annotations. Additionally, the trained models are less easily shareable than lexicons, which can hinder the reproducibility of research findings.

The advancement of Generative AI and large language models (LLMs) has inspired researchers to use these tools to derive text measures (*4–6*). Take the concept of political ideology (*7*), for example. Instead of training a machine learning model to classify a text as conservative or liberal, researchers can directly prompt an LLM to determine whether the text leans conservative or liberal, or even ask it to rate the political stance on a continuum. While LLM prompting offers greater flexibility and precision compared to traditional methods, the results remain sensitive to the design of the prompts, which can impact the reproducibility of findings (*8*, *9*). Over the years, researchers have sought text analysis methods that are valid, consistent, and reproducible for deriving text measures. In the era of LLMs, alongside LLM prompting, we ask: can we fully harness the potential of LLMs to achieve these goals?

Recent work has proposed an emerging area of representation learning aimed at better understanding how LLMs function internally, thereby enhancing the transparency of AI systems (*10*). Motivated by this, we introduce a new method, LLM-Measure, that leverages the internal hidden states of LLMs to generate text-based measures of interest. Take the concept of political ideology, for example. LLM-Measure consists of two stages: a probing stage and an inference stage. In the first probing stage, the goal is to learn an "ideology vector" that captures the direction of political stance. To do this, it first appends definitions of "conservative" and "liberal" to a reference text. Both versions of the text (conservative and liberal) are then passed through the LLM, which generates hidden states for each. The difference between the hidden states of the "conservative" and "liberal" versions is analyzed, and the most important components of this difference form the ideology vector. This vector serves as an axis representing the direction of political stance, pointing from conservative to liberal. In the second inference stage, LLM-Measure then assesses the political stance of a new text by comparing it to the ideology vector. The hidden states of the new input text are extracted by passing the text through the LLM. The method then projects these hidden states onto the ideology vector, with the resulting scalar value reflecting how strongly the input text aligns with conservative or liberal ideology.

We replicate three text-based concepts from prior social science research. The first measures the Fed's monetary stance at the macroeconomic level. The second evaluates company innovation at the microeconomic level, and the third assesses information overload in online reviews at the individual level. Our replication experiment, conducted using the open-source LLM LLaMA-2-Chat (13B) (*11*), demonstrates several advantages over existing text-based measurement methods. First, in terms of **validity**, the derived text concept measures exhibit high validity and significantly stronger statistical power compared to existing methods. Second, in terms of **consistency**, the measures are highly consistent across multiple runs and are not sensitive to variations in LLM prompt design. Third, in terms of **reproducibility**, our method leverages publicly available open-source LLMs, allowing researchers to openly share their prompts, which makes the results reproducible and supports more transparent quantitative social science research. Finally, our method is highly **efficient** because it does not rely on annotated data, making it particularly valuable in research scenarios where labelled data is difficult to obtain.

**Method**

In a typical quantitative social science research scenario involving text, a researcher compiles a dataset consisting of observations, where each observation includes a text input along with several covariates and/or a dependent variable (*1*). The goal of the text-based concept measurement is to derive a value for each text entry in the dataset that aligns with the specific concept the researcher aims to investigate. This value should reflect the underlying definition of the concept in the context of the research. In the following, we briefly describe the proposed LLM-Measure method.

*LLM hidden states*

LLM hidden states are the internal representations generated by LLMs as they process input sequences. Each token in the text sequence passes through multiple layers of the LLM, and at each layer, a hidden state is produced for every token. In autoregressive LLMs like GPT (*12*) and Llama (*11*), the hidden state of the last token captures information from the entire input sequence, as by the end of processing, the model has incorporated context from all previous tokens. As a result, the hidden state of the last token from the last layer is commonly used as the text embedding, representing the overall meaning of the input sequence. LLM hidden states are high-dimensional dense vectors, and their size varies depending on the model. For example, in Llama2, the hidden state of the last token has 5,120 dimensions.

*Probing stage: Obtain concept vector*

Most measurable concepts in social science and related fields have a directional nature with two opposing extremes, and their values can be measured along a continuum. In the probing stage, we aim to learn the direction of this continuum. For any given concept, we begin by defining two key points: the positive extreme, which represents the maximum value and is referred to as the "positive concept prompt," and the negative extreme, which represents the minimum value, referred to as the "negative concept prompt." For example, when measuring the Federal Reserve's monetary stance, the positive concept prompt might represent a hawkish view, while the low concept prompt would represent a dovish view (see Materials and Methods in supplementary materials).

Next, we compile a probing set, which consists of a small random sample of text taken from the entire sample. For each text in the probing set, we first append the positive concept prompt to the text and pass it through an LLM, extracting the corresponding hidden states from the last token of the last layer. Since the positive concept prompt emphasizes the positive side of the concept, the LLM is likely to focus on the positive aspects of that concept when generating the hidden states. We then repeat the process using the negative concept prompt to obtain the negative hidden states. Afterward, we compute the difference between the positive and negative hidden states by performing an element-wise subtraction of the two vectors. The resulting difference vector ideally encodes only the specific semantics of the concept, as other unrelated semantics are canceled out in the subtraction process.

We then apply Principal Component Analysis (PCA) to the difference vectors from the probing set. The largest principal component is considered the "**concept vector**," which ideally captures how the LLM internally represents the concept. In our three replication studies the concept vector derived from applying PCA shows a clear dominance in the first principal component (Fig. 1). This dominance suggests that the concept vector effectively captures the specific semantics of the concept.

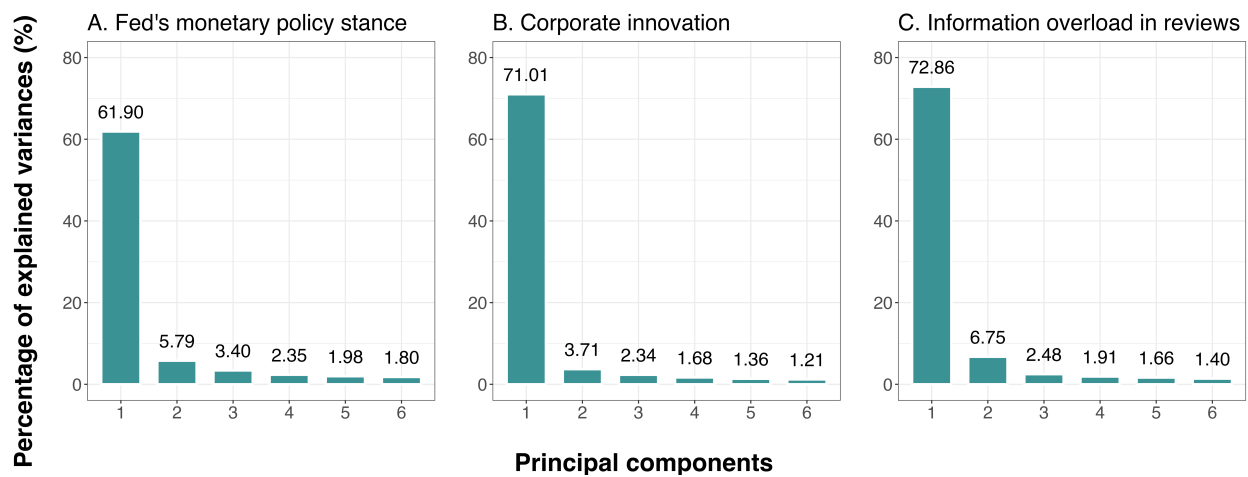

**Fig. 1. Percentage of variances explained by the principal components.** The largest principal component is considered as the concept vector.

*Inference stage: Obtain concept measures*

We then use the concept vector to determine the concept value for the entire text sample. To do this, we merge the positive and negative concept prompts into a single unified prompt, encouraging the LLM to focus on the specific semantics of the concept without emphasizing directionality. For each input text, we append the concept prompt and pass it through the LLM to generate the final hidden states. The concept value is determined by projecting the hidden state onto the concept vector. A large projection value indicates that the input text strongly reflects the concept being measured, while the sign of the projection shows the direction: a positive sign represents the positive side of the concept, and a negative sign represents the opposite (table S7).

Unlike the LLM prompting method, which directly asks an LLM to provide a score for a concept in a text, LLM-Measure analyzes the internal states of the LLM to understand how it encodes a particular concept. This allows for a more targeted and precise measurement of the concept within the text, based on how the LLM internally represents it.

**Experiments and Results**

To demonstrate the effectiveness and broad applicability of the LLM-Measure for text-based social science research, we replicate three text-based concepts that have been developed in prior literature. We also implement the LLM-prompting method for comparison (*5*). We validate our measures by examining their correlations with well-established dependent variables through ordinary least squares (OLS) regressions. We then compare the results to those obtained from other measurement approaches used in prior relevant studies.

| Dependent variable | Text Measure | Data | Method | Ref. | N | β | t |
|---|---|---|---|---|---|---|---|
| Panel A: Federal Reserve's monetary stance | | | | | | | |
| 10-year US Treasury yield | Hawkishness of monetary policy stance | FOMC Minutes | RoBERTa | (*13*) | 62 | 0.252 | 2.949 |
| | | | LLM-prompting | (*5*) | 62 | 0.159 | 1.780 |
| | | | LLM-Measure | Ours | 62 | 0.382 | 4.957 |
| Panel B: Innovation level within a company | | | | | | | |
| R&D spending | Corporate innovation | Earnings transcripts | Dictionary | (*14*) | 5,237 | 0.010 | 3.193 |
| | | | LLM-prompting | (*5*) | 5,237 | 0.035 | 11.408 |
| | | | LLM-Measure | Ours | 5,237 | 0.031 | 10.152 |
| Panel C: Information overload in online reviews | | | | | | | |
| Review helpfulness | Information overload | Online reviews | LDA | (*15*) | 10,000 | -0.018 | -2.553 |
| | | | LLM-prompting | (*5*) | 10,000 | 0.024 | 3.095 |
| | | | LLM-Measure | Ours | 10,000 | -0.044 | -5.696 |

**Table 1. Summary of replication study results.** Each row corresponds to an OLS regression where the independent variable is measured using either the proposed LLM-Measure, LLM-prompting or established methods from previous studies. *N* refers to the number of observations, and *β* represents the effect size, indicating the strength of the association between the text measure and the dependent variable. We also report *t-statistics* to indicate the statistical significance of the relationship. All independent variables are standardized to allow for the comparison of effect sizes (see the supplementary materials for more details on the replication studies).

*Study 1: Federal Reserve's monetary stance*

The first macro-economic level concept is the Fed's monetary policy stance. This concept is critical in social science research as it shapes economic conditions, financial markets, and policymaking, influencing both domestic and global economies (*16*). Furthermore, recent studies have utilized textual analysis of the Federal Open Market Committee (FOMC) minutes to better understand shifts in the Fed's policy stance, highlighting the role of language in shaping monetary policy decisions (*17*).

Prior work by Shah *et al.* (*13*) benchmarks various supervised machine learning models for classifying sentences in the FOMC minutes as hawkish or dovish. Their best-performing model is fine-tuning a RoBERTa-large model (*18*). Additionally, they find a significant positive correlation between a hawkish stance and higher bond yields, demonstrating the validity of the text-based monetary stance measure. In this replication study, we apply LLM-Measure to derive the Fed's monetary policy stance and examine its correlation with U.S. Treasury yields.

*Study 2: Corporate innovation level*

The second micro-economic level concept is a company's level of innovation. Innovation is a key driver of competitive advantage, productivity, and long-term growth, making it a central focus in social science research on firm performance and economic development (*19*). Understanding a company's innovation level helps researchers analyze its potential for technological advancement and its ability to respond to market changes (*20*).

To measure corporate innovation, Li *et al.* (*14*) introduce a keyword-based method by developing an innovation-related lexicon and applying it to companies' earnings conference call transcripts. They validate their innovation measure by correlating it with the firm's R&D spending, finding a significant positive correlation. In this replication study, we apply LLM-Measure to derive corporate innovation and examine its correlation with R&D spending.

*Study 3: Information overload in online reviews*

The third individual-level concept is the information load on consumer reviews. Information overload in online reviews can significantly impact consumer decision-making, as too much information can lead to cognitive fatigue, reducing the ability to make informed choices (*21*). Studying this concept is crucial in social science research to understand how consumers process information in the digital age and how it affects purchasing behavior and satisfaction.

Prior work by Ghose *et al.* (*15*) proposes an unsupervised machine learning method to measure information overload in online reviews. Specifically, they apply Latent Dirichlet Allocation (LDA) (*22*) to review datasets and define information overload as the entropy of a review's topic distribution. They validate this measure by demonstrating a significant negative correlation between information overload and the rated helpfulness of reviews (*15*), indicating that more complex and noisier reviews tend to be perceived as less helpful due to cognitive overload. In this replication study, using an online hotel review dataset sourced from TripAdvisor.com, we apply LLM-Measure to derive the information overload measure and its relationship with perceived review helpfulness.

*Results*

The replication results lead to several key findings (Table 1 and tables S1 to S3). First, across all three studies, the text measures derived from LLM-Measure show significantly larger effect sizes and stronger statistical significance compared to methods used in prior studies. For instance, in study 2, the relationship between a firm's corporate innovation and its R&D spending is stronger when using LLM-Measure ($\beta = 0.031$, $t = 10.152$) compared to the dictionary-based approach ($\beta = 0.010$, $t = 3.193$), indicating greater external validity and economic significance. Second, while the LLM-prompting method yields slightly higher effect sizes and t-statistics than LLM-Measure in study 2, it is important to note that LLM-prompting is highly sensitive to the prompts used, raising concerns about the reliability and reproducibility of the results. In study 3, the LLM-prompting method even produces an unexpected and contradictory relationship between information overload and review helpfulness, likely due to measurement errors, as the measure of information overload in online reviews requires more than semantic interpretation beyond the capability of LLM prompting. Finally, compared to dictionary-based and machine learning-based methods, LLM-Measure does not rely on human data annotations. Overall, this replication experiment demonstrates that LLM-Measure is adaptable to various research contexts and can be applied to a wide range of text measures, making it a valuable tool for analyzing diverse social science concepts.

**LLM Prompt Sensitivity Test**

Consistency and reproducibility are considered essential for ensuring the repeatability of study outcomes and the progression of scholarly work (*23*, *24*). However, one major criticism of the LLM-based method is that the values derived can vary significantly depending on how the prompt is phrased, raising concerns about consistency and reliability (*8*, *9*). In contrast, our proposed method, which relies on PCA analysis of the internal LLM hidden states, is much less sensitive to changes in the prompt. To demonstrate this, we conducted two tests: in the first, we modify the concept definition by altering the positive and negative concept prompts; in the second, we change the instructions used in the prompt. We then examined the correlation between the concept measures derived using the original prompts and those derived from rephrased prompts.

The results show that LLM-prompting may lack consistence across different prompt variations (Fig. 2 and fig. S1). For example, for study 3 of measuring information overload in online reviews, changing the concept definition results in only a 0.539 correlation between the measures, whereas the correlation for LLM-Measure is as high as 0.958 (Fig. 2C). This sensitivity to prompt variation in LLM-prompting raises concerns about the reproducibility and validity of research findings. In addition, measures derived from LLM-Measure remain highly consistent across different sizes of the probing set (fig. S2). Overall, LLM-Measure significantly improves measurement consistency when using LLMs.

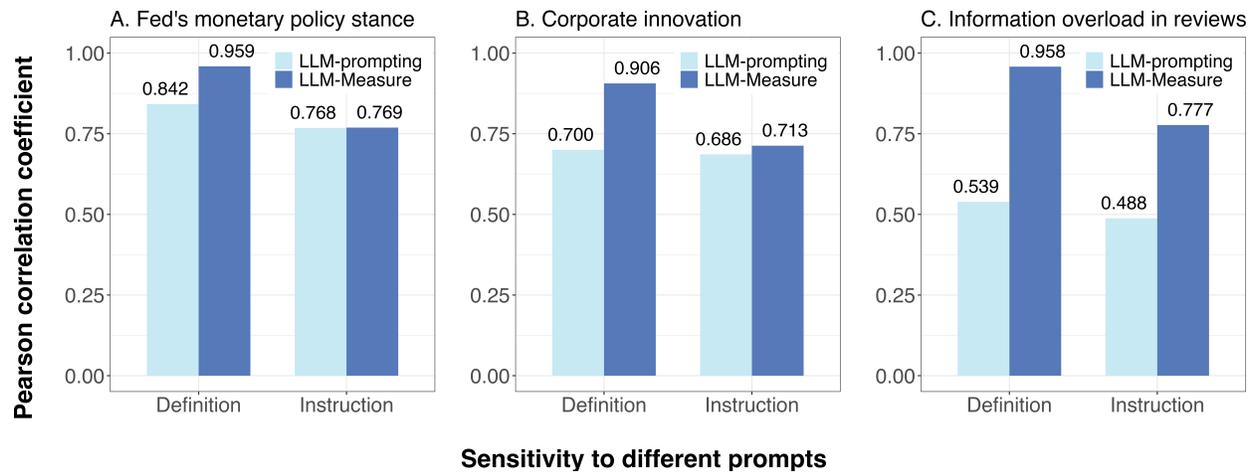

**Fig. 2. Sensitivity analysis of LLM-prompting and LLM-Measure to the prompt format.**

**Discussion and Conclusion**

In this paper, we propose LLM-Measure, a new method for deriving text-based measures of concepts in social science research context. Looking back at the evolution of text analysis tools for social scientists—from keyword and rule-based methods to machine learning models, and more recently, to LLM prompting—we observe a consistent pursuit of approaches that provide high validity, consistency, and reproducibility, while reducing the need for human intervention. LLMs, a groundbreaking advancement in understanding complex language, offer researchers new opportunities to leverage large text datasets to create concepts, test hypotheses, and validate existing ones, helping to understand societal and economic phenomena and their policy implications. In this paper, through three replication studies, we demonstrate that measures derived from an LLM's internal hidden states achieve greater validity, stability, and efficiency compared to those generated through direct prompting, a method increasingly adopted in social science research (*4–6*). Moreover, LLM-Measure does not require model training and relies solely on open-source LLMs, allowing researchers to ensure the reproducibility of their work by simply sharing the prompts they used, without needing to release training data or trained models.

Researchers should also be aware of potential risks associated with the LLM-generated scores. For instance, LLMs may exhibit certain biases, with responses being more liberal than those of the general population and reflecting the views of younger, more educated individuals (*25*). That being said, opening the black box and investigating the internal hidden states may improve the transparency and give researchers greater controls. We believe this work is timely for social science researchers, offering an effective and efficient way to develop text-based measures in their empirical studies, while also aligning with the growing emphasis on improving reproducibility in scientific research (*23*, *24*, *26*).

**References and Notes**

1. M. Gentzkow, B. Kelly, M. Taddy, Text as data. *Journal of Economic Literature* **57**, 535–574 (2019).

2. M. E. Roberts, B. M. Stewart, E. M. Airoldi, A model of text for experimentation in the social sciences. *Journal of the American Statistical Association* **111**, 988–1003 (2016).

3. J. Wilkerson, A. Casas, Large-scale computerized text analysis in political science: Opportunities and challenges. *Annual Review of Political Science* **20**, 529–544 (2017).

4. S. Rathje, D.-M. Mirea, I. Sucholutsky, R. Marjieh, C. E. Robertson, J. J. Van Bavel, GPT is an effective tool for multilingual psychological text analysis. *Proceedings of the National Academy of Sciences* **121**, e2308950121 (2024).

5. G. Le Mens, B. Kovács, M. T. Hannan, G. Pros, Uncovering the semantics of concepts using GPT-4. *Proceedings of the National Academy of Sciences* **120**, e2309350120 (2023).

6. F. Gilardi, M. Alizadeh, M. Kubli, ChatGPT outperforms crowd workers for text-annotation tasks. *Proceedings of the National Academy of Sciences* **120**, e2305016120 (2023).

7. M. Gentzkow, J. M. Shapiro, M. Taddy, Measuring group differences in high-dimensional choices: method and application to congressional speech. *Econometrica* **87**, 1307–1340 (2019).

8. C. A. Bail, Can Generative AI improve social science? *Proceedings of the National Academy of Sciences* **121**, e2314021121 (2024).

9. M. Sclar, Y. Choi, Y. Tsvetkov, A. Suhr, Quantifying Language Models' Sensitivity to Spurious Features in Prompt Design or: How I learned to start worrying about prompt formatting. *arXiv:2310.11324* (2023).

10. A. Zou, L. Phan, S. Chen, J. Campbell, P. Guo, R. Ren, A. Pan, X. Yin, M. Mazeika, A.-K. Dombrowski, Representation engineering: A top-down approach to ai transparency. *arXiv:2310.01405* (2023).

11. H. Touvron, L. Martin, K. Stone, P. Albert, A. Almahairi, Y. Babaei, N. Bashlykov, S. Batra, P. Bhargava, S. Bhosale, Llama 2: Open foundation and fine-tuned chat models. *arXiv:2307.09288* (2023).

12. T. Brown, B. Mann, N. Ryder, M. Subbiah, J. D. Kaplan, P. Dhariwal, A. Neelakantan, P. Shyam, G. Sastry, A. Askell, S. Agarwal, A. Herbert-Voss, G. Krueger, T. Henighan, R. Child, A. Ramesh, D. Ziegler, J. Wu, C. Winter, C. Hesse, M. Chen, E. Sigler, M. Litwin, S. Gray, B. Chess, J. Clark, C. Berner, S. McCandlish, A. Radford, I. Sutskever, D. Amodei, "Language Models are Few-Shot Learners" in *Proceedings of the 34th International Conference on Neural Information Processing Systems* (Curran Associates Inc., 2020), pp. 1877–1901.

13. A. Shah, S. Paturi, S. Chava, "Trillion Dollar Words: A New Financial Dataset, Task & Market Analysis" in *Proceedings of the 61st Annual Meeting of the Association for Computational Linguistics* (Association for Computational Linguistics, 2023), pp. 6664–6679.

14. K. Li, F. Mai, R. Shen, X. Yan, Measuring Corporate Culture Using Machine Learning. *The Review of Financial Studies* **34**, 3265–3315 (2021).

15. A. Ghose, P. G. Ipeirotis, B. Li, Modeling consumer footprints on search engines: An interplay with social media. *Management Science* **65**, 1363–1385 (2019).

16. B. S. Bernanke, A. S. Blinder, The Federal Funds Rate and the Channels of Monetary Transmission. *American Economic Review* **82**, 901–921 (1992).

17. S. Hansen, M. McMahon, A. Prat, Transparency and deliberation within the FOMC: A computational linguistics approach. *The Quarterly Journal of Economics* **133**, 801–870 (2018).

18. Y. Liu, M. Ott, N. Goyal, J. Du, M. Joshi, D. Chen, O. Levy, M. Lewis, L. Zettlemoyer, V. Stoyanov, Roberta: A robustly optimized bert pretraining approach. *arXiv:1907.11692* (2019).

19. J. A. Schumpeter, *Capitalism, Socialism and Democracy* (routledge, 2013).

20. P. Aghion, P. W. Howitt, *The Economics of Growth* (MIT press, 2008).

21. J. Jacoby, Perspectives on information overload. *Journal of consumer research* **10**, 432–435 (1984).

22. D. M. Blei, A. Y. Ng, M. I. Jordan, Latent dirichlet allocation. *Journal of machine Learning research* **3**, 993–1022 (2003).

23. B. A. Nosek, G. Alter, G. C. Banks, D. Borsboom, S. D. Bowman, S. J. Breckler, S. Buck, C. D. Chambers, G. Chin, G. Christensen, Promoting an open research culture. *Science* **348**, 1422–1425 (2015).

24. V. Stodden, M. McNutt, D. H. Bailey, E. Deelman, Y. Gil, B. Hanson, M. A. Heroux, J. P. Ioannidis, M. Taufer, Enhancing reproducibility for computational methods. *Science* **354**, 1240–1241 (2016).

25. S. Santurkar, E. Durmus, F. Ladhak, C. Lee, P. Liang, T. Hashimoto, "Whose opinions do language models reflect?" in *Proceedings of the 40th International Conference on Machine Learning* (PMLR, 2023), pp. 29971–30004.

26. I. Grossmann, M. Feinberg, D. C. Parker, N. A. Christakis, P. E. Tetlock, W. A. Cunningham, AI and the transformation of social science research. *Science* **380**, 1108–1109 (2023).

**Data and materials availability:** Code and data for reproducing the reported results are available in https://github.com/hduanac/LLM-Measure

**Supplementary Materials**
Materials and Methods
Supplementary Text
Figs. S1 to S2
Tables S1 to S7
References

# Supplementary Materials for

## LLM-Measure: Generating Valid, Consistent, and Reproducible Text-Based Measures for Social Science Research

Yi Yang[1]*, Hanyu Duan[1], Jiaxin Liu[1], Kar Yan Tam[1]

*Corresponding author. Email: imyiyang@ust.hk

**The PDF file includes:**

>Materials and Methods
>Supplementary Text
>Figs. S1 to S2
>Tables S1 to S7
>References

**Materials and Methods**

**LLM-Measure implementation details.** We include the implementation details for LLM-Measure to support reproducibility.

    **Large language model.** LLM-Measure can accommodate any open-source Transformer-based LLMs. For this study, we choose LLaMA-2-Chat (13B) (*11*). We download the model following the instructions from Meta AI [1] and implement it using the *transformers* library. [2]

    **Handling long text.** We apply chunking strategies in case the input sequence is too long, which might lead to decreased processing efficiency or exceed the language model's maximum input limit (4,096 tokens for LLaMA-2). More specifically, for text longer than 1,000 tokens, we first divide it into chunks of up to 1,000 tokens each and then aggregate their associated hidden states via average pooling.

    **Principal component analysis (PCA).** During the probing stage, to ensure PCA functions properly, we standardize all the difference vectors to have a mean of zero and a standard deviation of one before applying PCA to them. Similarly, at the inference stage, we also

---

[1] https://llama.meta.com/llama-downloads/
[2] https://github.com/huggingface/transformers/

standardize each hidden state (using the parameters obtained from the probing stage) before projecting it onto the concept vector. Both PCA and standardization are implemented using the *scikit-learn* library in Python. [3]

**Probing set.** The probing set is created by randomly picking a small portion of samples from the full dataset. Our main results are based on 64 samples. This number is sufficient to ensure converged measures, and further increasing the sample size does not lead to significant changes in the outcomes (Fig. S2).

**Direction of the concept vector.** To ensure concept vectors consistently point from the negative end to the positive end across our replication studies, for each text measure, we examine examples in the probing set that have extreme projection values (see Table S7, the review study, for an example). If the examples with the top (bottom) projection values align with the positive (negative) side of the concept, it confirms that the concept vector is correctly oriented. Otherwise, we flip the direction of the concept vector by multiplying it by -1.

**Supplementary Text**

**LLM-Measure prompts.** Here, we list the prompts used by LLM-Measure in our experiments.

***Study 1: Federal Reserve's monetary stance:***

- Positive concept prompt: *"[INST] ≪SYS≫ Pretend you are a financial analyst. Your task is to analyze the monetary policy stance of a text snippet from the Federal Open Market Committee (FOMC) document. Particularly, you should focus on what traits in the text snippet suggesting that this text is being hawkish. hawkish is defined by more restraint in growing various measures of money supply like M0, M1, M2, more concerned about risks of excess money supply expansion fueling inflation in the medium-term, maintaining strong controls on the monetary base to safeguard purchasing power, push up short-term borrowing costs and lowers asset valuations like stocks or housing. ≪/SYS≫ what traits suggest that the following text snippet (delimited by triple quotes) was being hawkish? Text snippet: ```Input Sequence```. [/INST] The traits making it sound like hawkish include"*

- Negative concept prompt: *"[INST] ≪SYS≫ Pretend you are a financial analyst. Your task is to analyze the monetary policy stance of a text snippet from the Federal Open*

---
[3] https://github.com/scikit-learn/scikit-learn/

*Market Committee (FOMC) document. Particularly, you should focus on what traits in the text snippet suggesting that this text is being dovish. dovish is defined by more willing to grow the money supply at a robust pace through measures like quantitative easing, less concerned about risks of too much money supply expansion fueling inflation in the short-run, giving a boost to aggregate demand, bank lending, and overall economic activity by ensuring ample liquidity in the financial system. ≪/SYS≫ what traits suggest that the following text snippet (delimited by triple quotes) was being dovish? Text snippet: ```Input Sequence```. [/INST] The traits making it sound like dovish include"*

- Unified concept prompt: *"[INST] ≪SYS≫ Pretend you are a financial analyst. Your task is to analyze the monetary policy stance of a text snippet from the Federal Open Market Committee (FOMC) document. Particularly, you should focus on what traits in the text snippet suggesting that this text is being hawkish or being dovish. hawkish is defined by more restraint in growing various measures of money supply like M0, M1, M2, more concerned about risks of excess money supply expansion fueling inflation in the medium-term, maintaining strong controls on the monetary base to safeguard purchasing power, push up short-term borrowing costs and lowers asset valuations like stocks or housing. dovish is defined by more willing to grow the money supply at a robust pace through measures like quantitative easing, less concerned about risks of too much money supply expansion fueling inflation in the short-run, giving a boost to aggregate demand, bank lending, and overall economic activity by ensuring ample liquidity in the financial system. ≪/SYS≫ what traits suggest that the following text snippet (delimited by triple quotes) was being hawkish or being dovish? Text snippet: ```Input Sequence```. [/INST] The traits contribute to the monetary policy stance include"*

**Study 2: Corporate innovation level:**

- Positive concept prompt: *"[INST] ≪SYS≫ Pretend you are a financial analyst. Your task is to analyze the firm's innovation of a text snippet from the firm's earning conference call. Particularly, you should focus on what traits in the text snippet suggesting that this text is being innovative. innovative is defined by the creation or implementation of new processes, products or services by a firm to advance, compete and differentiate themselves in the market. ≪/SYS≫ what traits suggest that the following text snippet* (delimited by triple quotes) was being innovative? Text snippet: ```Input Sequence```. [/INST] The traits making it sound like innovative include"

- Negative concept prompt: *"[INST] ≪SYS≫ Pretend you are a financial analyst. Your task is to analyze the firm's innovation of a text snippet from the firm's earning conference call. Particularly, you should focus on what traits in the text snippet suggesting that this text is being conservative. conservative is defined by a reluctance to embrace bold innovations, digital disruption or make substantial deviations from tried-and-true products, processes and strategies. ≪/SYS≫ what traits suggest that the following text snippet (delimited by triple quotes) was being conservative? Text snippet: ```Input Sequence```. [/INST] The traits making it sound like conservative include"*

- Unified concept prompt: *"[INST] ≪SYS≫ Pretend you are a financial analyst. Your task is to analyze the firm's innovation of a text snippet from the firm's earning conference call. Particularly, you should focus on what traits in the text snippet suggesting that this text is being innovative or being conservative. innovative is defined by the creation or implementation of new processes, products or services by a firm to advance, compete and differentiate themselves in the market. conservative is defined by a reluctance to embrace bold innovations, digital disruption or make substantial deviations from tried-and-true products, processes and strategies. ≪/SYS≫ what traits suggest that the following text snippet (delimited by triple quotes) was being innovative or being conservative? Text snippet: ```Input Sequence```. [/INST] The traits contribute to the firm's innovation include"*

***Study 3: Information overload in online reviews:***

- Positive concept prompt: *"[INST] ≪SYS≫ Pretend you are a linguist. Your task is to analyze the content complexity of a customer review about hotels. Particularly, you should focus on the characteristics regarding the review content that may make the review complex. More formally, a complex review is characterized by comprehensive and detailed reviews that encompass a wide range of aspects, incorporate diverse topics, and provide nuanced insights into the hotel experience. ≪/SYS≫ What characteristics regarding the content of the following customer review about hotels (delimited by triple quotes), do you think, may contribute to a complex information complexity? Review text: ```Input Sequence```. [/INST] The characteristics making the review complex include"*

- Negative concept prompt: *"[INST] ≪SYS≫ Pretend you are a linguist. Your task is to analyze the content complexity of a customer review about hotels. Particularly, you should focus on the characteristics regarding the review content that may make the review concise. More formally, a concise review is characterized by simplistic and superficial reviews that lack depth, focus on narrow topics, and offer limited analysis of the hotel experience. ≪/SYS≫ What characteristics regarding the content of the following customer review about hotels (delimited by triple quotes), do you think, may contribute to a concise information complexity? Review text: ```Input Sequence```. [/INST] The characteristics making the review concise include"*

- Unified concept prompt: *"[INST] ≪SYS≫ Pretend you are a linguist. Your task is to analyze the content complexity of a customer review about hotels. Particularly, you should focus on the characteristics regarding the review content that may make the review either complex or concise. More formally, a complex review is characterized by comprehensive and detailed reviews that encompass a wide range of aspects, incorporate diverse topics, and provide nuanced insights into the hotel experience. A concise review is characterized by simplistic and superficial reviews that lack depth, focus on narrow topics, and offer limited analysis of the hotel experience. ≪/SYS≫ What characteristics regarding the content of the following customer review about hotels (delimited by triple quotes), do you think, may contribute to the content complexity of this review, making it either complex or concise? Review text: ```Input Sequence```. [/INST] The characteristics contributing to its content complexity include"*

**Regression details.** Here, we describe the regression details in our three replication studies.

**Correlation between the Fed's monetary policy stance and U.S. Treasury yield.** We estimate the U.S. Treasury yield by running the ordinary least squares (OLS) regression provided in the equation below:

$$Yield = \beta_0 + \beta_1 PolicyStance + \epsilon$$

The dependent variable, $Yield$, represents the U.S. Treasury yield for 10-year, 1-year, or 3-month maturities. The independent variable, $PolicyStance$, is defined as the hawkishness of the Fed's monetary policy stance and measured using RoBERTa (*18*), LLM-prompting, and our LLM-Measure approach, based on the FOMC press conference transcripts collected and processed by Shah *et al.* (*13*). The dataset used in our experiments includes 62 press conference documents in total from April 27, 2011, to October 15, 2022, with each document averaging 81 sentences. We obtain the document-level measure of hawkishness by averaging the sentence-level values. The regression results appear in Table S1, and the descriptive statistics of the main variables are presented in Table S4.

**Correlation between corporate innovation and R&D spending.** We estimate the OLS model shown in the equation below:

$$R\&Dspending = \beta_0 + \beta_1 Innovation + \beta_2 Size + \epsilon$$

The dependent variable, $R\&Dspending$, is coded as the firm's yearly R&D expenses normalized by the total assets. We control for firm size, $Size$, defined as the natural logarithm of the firm's total assets. The independent variable, $Innovation$, represents the corporate cultural value of innovation and is measured from the question-and-answer (QA) section of firm earnings call transcripts, using the Dictionary method (*14*), LLM-prompting, and our LLM-Measure. The earnings call transcripts are collected from Seeking Alpha [4] and cover a period from 2016 to 2020 and include 2101 unique companies. Given that the earnings call data is quarterly, we use average pooling to aggregate the quarterly innovation measures into an annual figure, and the final firm-year measures are based on 3-year moving averages of these annual values (*14*). The regression results and the descriptive statistics of the main variables used in the regression analysis are shown in Table S2 and Table S5, respectively.

**Correlation between information overload and helpfulness of online reviews.** We compile a dataset containing 10K online hotel reviews written in English based on an open-source dataset [5] sourced from TripAdvisor.com. We run the OLS model illustrated in the equation below:

$$\log(Helpfulness + 1) = \beta_0 + \beta_1 InformationOverload + \beta_2 Sentiment + \beta_3 \log(Words) + \epsilon$$

Review helpfulness, labeled as $Helpfulness$, is defined by the number of votes a review receives. Due to the highly skewed vote distribution, where most reviews receive no votes, we take the natural logarithm of the vote count in this OLS model (for robustness check, we estimate another OLS model without log transformation, and we report both results in Table S3). The independent variable, $InformationOverload$, is measured by LDA (*22*), LLM-prompting, and our LLM-Measure. We include two control variables: 1) $Sentiment$, a binary score derived from the review rating, set to 0 for ratings of three stars or fewer and 1 for ratings of four or five stars; and 2) $Words$, the number of words in the review. The descriptive statistics for the main variables used in the regression analysis are presented in Table S6.

---

[4] https://seekingalpha.com
[5] https://www.cs.cmu.edu/~jiweil/html/hotel-review.html

**Baseline methods.** Here, we describe the baseline methods applied in our replication studies.

**RoBERTa.** We use the measures released by Shah *et al.* (*13*), who measure the hawkishness of the Fed's monetary policy stance based on counting the number of hawkish (dovish) sentences in the FOMC press conference document, using the equation below:

$$PolicyStance = \frac{\#Hawkish - \#Dovish}{\#Total}$$

Here, $\#Hawkish$ ($\#Dovish$) denotes the number of hawkish (dovish) sentences predicted by a finetuned RoBERTa model using manually annotated data. $\#Total$ is the total number of sentences in the document.

**LDA.** Following Ghose *et al.* (*15*), we use the latent Dirichlet allocation (LDA) (*22*) model to extract topic features from reviews. We measure the information overload of a review as the entropy (Shannon index) of its topic features:

$$InformationOverload_i = -\sum_{j=1}^{J} Topic_{ij} \log(Topic_{ij})$$

Here $Topic_{ij}$ denotes the $j$-th topic's probability of the $i$-th review. The higher the entropy, the heavier the information load of the review. We implement LDA using the *LatentDirichletAllocation* library in *scikit-learn* [6]. Before fitting the model, we perform stemming and remove stop words, punctuations, and special characters. We exclude terms that appear in more than 50% of the reviews or in fewer than 5 reviews. We set the number of topics to $J = 25$ and the learning rate decay to 0.1.

**LLM-prompting.** We implement LLM-prompting using LLaMA-2-Chat (13B). We craft the prompts following Mens *et al.* (*5*) and set the default temperature to 1. We apply a chunking-and-pooling (average) strategy to handle input data that exceeds 1,000 tokens. The prompts used in each study are listed below.

---

[6] https://scikit-learn.org/stable/modules/generated/sklearn.decomposition.LatentDirichletAllocation.html

**Implementation of LLM-prompting method.** We provide the prompts used for LLM-prompting in our three studies.

*Study 1: Federal Reserve's monetary stance:*

- Prompt: *"[INST] <<SYS>> Pretend you are a financial analyst. Your task is to analyze the monetary policy stance of a text snippet from the Federal Open Market Committee (FOMC) document. Particularly, you should focus on what traits in the text snippet suggesting that this text is being hawkish or dovish? Hawkish is defined by more restraint in growing various measures of money supply like M0, M1, M2, more concerned about risks of excess money supply expansion fueling inflation in the medium-term, maintaining strong controls on the monetary base to safeguard purchasing power, push up short-term borrowing costs and lowers asset valuations like stocks or housing. Dovish is defined by more willing to grow the money supply at a robust pace through measures like quantitative easing, less concerned about risks of too much money supply expansion fueling inflation in the short-run, giving a boost to aggregate demand, bank lending, and overall economic activity by ensuring ample liquidity in the financial system. <</SYS>> Here's the text snippet:```Input Sequence```. [/INST] How hawkish or dovish is this text snippet of the monetary policy stance? Provide your response as a score between 0 and 100 where 0 means dovish and 100 means hawkish. The higher the score, the more hawkish the text is. Directly output the score without giving the explanation. ###SCORE:"*

- Prompt (definition rephrased): *"[INST] <<SYS>> Pretend you are a financial analyst. Your task is to analyze the monetary policy stance of a text snippet from the Federal Open Market Committee (FOMC) document. Particularly, you should focus on what traits in the text snippet suggesting that this text is being hawkish or dovish? Hawkish is defined by greater restraint in expanding money supply measures like M0, M1, and M2; increased concern over medium-term inflation risks from excess money supply; stringent controls on the monetary base to protect purchasing power; higher short-term borrowing costs; reduced asset valuations, including stocks and housing. Dovish is defined by* *greater willingness to expand the money supply rapidly through measures like quantitative easing; reduced concern about short-run inflation risks from excessive money supply growth; stimulation of aggregate demand, bank lending, and overall economic activity by providing ample liquidity in the financial system. <</SYS>> Here's the text snippet:```Input Sequence```. [/INST] How hawkish or dovish is this text snippet of the monetary policy stance? Provide your response as a score between 0 and 100 where 0 means dovish and 100 means hawkish. The higher the score, the more hawkish the text is. Directly output the score without giving the explanation. ###SCORE:"*

- Prompt (instruction rephrased): *"[INST] <<SYS>> Imagine you are a financial analyst tasked with analyzing the monetary policy stance of a sentence from the Federal Open Market Committee (FOMC) document. Hawkish sentence is characterized by more restraint in growing various measures of money supply like M0, M1, M2, more concerned about risks of excess money supply expansion fueling inflation in the medium-term, maintaining strong controls on the monetary base to safeguard purchasing power, push up short-term borrowing costs and lowers asset valuations like stocks or housing, whereas dovish sentence is defined as more willing to grow the money supply at a robust pace through measures like quantitative easing, less concerned about risks of too much money supply expansion fueling inflation in the short-run, giving a boost to aggregate demand, bank lending, and overall economic activity by ensuring ample liquidity in the financial system. <</SYS>> Sentence:```Input Sequence```. [/INST] How hawkish the sentence is? Assign a score between 0 and 100, where 0 indicates very dovish and 100 indicates very hawkish. The higher the score, the greater the hawkish of the sentence. Output only the score without giving the explanation. ###SCORE:"*

**Study 2: Corporate innovation level:**

- Prompt: *"[INST] <<SYS>> Pretend you are a financial analyst. Your task is to analyze the firm's innovation of a text snippet from the firm's earning conference call. Particularly, you should focus on what traits in the text snippet suggesting that this text is being innovative or conservative? Innovative is defined by the creation or implementation of new processes, products or services by a firm to advance, compete and differentiate themselves in the market. Conservative is defined by a reluctance to embrace bold innovations, digital disruption or make substantial deviations from tried-* *and-true products, processes and strategies. <</SYS>> Here's the text snippet:```Input Sequence```. [/INST] How innovative or conservative is this text snippet of the firm's innovation? Provide your response as a score between 0 and 100 where 0 means conservative and 100 means innovative. The higher the score, the more innovative the text is. Directly output the score without giving the explanation. ###SCORE:"*

- Prompt (definition rephrased): *"[INST] <<SYS>> Pretend you are a financial analyst. Your task is to analyze the firm's innovation of a text snippet from the firm's earning conference call. Particularly, you should focus on what traits in the text snippet suggesting that this text is being innovative or conservative? Innovative is defined by developing or implementing innovative processes, products, or services to advance, compete, and differentiate in the market. Conservative is defined by reluctance to adopt bold innovations or digital disruption, and to deviate significantly from established products, processes, and strategies. <</SYS>> Here's the text snippet:```Input Sequence```. [/INST] How innovative or conservative is this text snippet of the firm's innovation? Provide your response as a score between 0 and 100 where 0 means conservative and 100 means innovative. The higher the score, the more innovative the text is. Directly output the score without giving the explanation. ###SCORE:"*

- Prompt (instruction rephrased): *"[INST] <<SYS>> Imagine you are a financial analyst tasked with analyzing the firm's innovation of a sentence from the firm's earning conference call. Innovative sentence is characterized by the creation or implementation of new processes, products or services by a firm to advance, compete and differentiate themselves in the market, whereas conservative sentence is defined as a reluctance to embrace bold innovations, digital disruption or make substantial deviations from tried-and-true products, processes and strategies. <</SYS>> Sentence:```Input Sequence```. [/INST] How innovative the sentence is? Assign a score between 0 and 100, where 0 indicates very conservative and 100 indicates very innovative. The higher the score, the greater the innovative of the sentence. Output only the score without giving the explanation. ###SCORE:"*

**Study 3: Information overload in online reviews:**
- Prompt: *"[INST] <<SYS>> Pretend you are a linguist. Your task is to evaluate the content complexity of a customer review about hotels. By definition, complex reviews are* *comprehensive and detailed reviews that encompass a wide range of aspects, incorporate diverse topics, and provide nuanced insights into the hotel experience, while concise reviews are simplistic and superficial reviews that lack depth, focus on narrow topics, and offer limited analysis of the hotel experience. <</SYS>> Review text: ```Input Sequence```. [/INST] How complex is this review? Provide your response as a score between 0 and 100 where 0 means extremely concise and 100 means extremely complex. The higher the score, the more complex the review is. Directly output the score without giving the explanation. ###SCORE:"*

- Prompt (definition rephrased): *"[INST] <<SYS>> Pretend you are a linguist. Your task is to evaluate the content complexity of a customer review about hotels. By definition, complex reviews are extensive and intricate assessments that cover a broad spectrum of elements, incorporate varied subjects, and offer subtle perspectives on the hotel stay, while concise reviews are straightforward and surface-level evaluations that lack profundity, concentrate on specific subjects, and provide restricted analysis of the hotel stay. <</SYS>> Review text: ```Input Sequence```. [/INST] How complex is this review? Provide your response as a score between 0 and 100 where 0 means extremely concise and 100 means extremely complex. The higher the score, the more complex the review is. Directly output the score without giving the explanation. ###SCORE:"*

- Prompt (instruction rephrased): *"[INST] <<SYS>> Imagine you are a linguist tasked with assessing the content complexity of a hotel customer review. complex reviews are characterized by comprehensive and detailed reviews that encompass a wide range of aspects, incorporate diverse topics, and provide nuanced insights into the hotel experience, whereas concise reviews are defined as simplistic and superficial reviews that lack depth, focus on narrow topics, and offer limited analysis of the hotel experience. <</SYS>> Hotel review: ```Input Sequence```. [/INST] How complex the review is? Assign a score between 0 and 100, where 0 indicates very concise and 100 indicates very complex. The higher the score, the greater the complexity of the review. Output only the score without any explanation. ###SCORE:"*

**Prompts for sensitivity test.** We provide the prompts used for LLM sensitivity test.

***Study 1: Federal Reserve's monetary stance:***

- Positive concept prompt (definition rephrased): *"[INST] ≪SYS≫ Pretend you are a financial analyst. Your task is to analyze the monetary policy stance of a text snippet from the Federal Open Market Committee (FOMC) document. Particularly, you should focus on what traits in the text snippet suggesting that this text is being hawkish. hawkish is defined by more constraint in expanding diverse measures of money supply like M0, M1, M2, more worried about risks stemming from excess money supply contributing to inflation in the medium-term, implementing robust supervision on the monetary foundation served to shield purchasing power, elevate short-term borrowing costs and decrease valuations for assets such as stocks or housing. ≪/SYS≫ what traits suggest that the following text snippet (delimited by triple quotes) was being hawkish? Text snippet: ```Input Sequence```. [/INST] The traits making it sound like hawkish include"*

- Negative concept prompt (definition rephrased): *"[INST] ≪SYS≫ Pretend you are a financial analyst. Your task is to analyze the monetary policy stance of a text snippet from the Federal Open Market Committee (FOMC) document. Particularly, you should focus on what traits in the text snippet suggesting that this text is being dovish. dovish is defined by more amenable to expanding the money supply at a vigorous pace through means such as quantitative easing, less worried about risks stemming from excess money supply contributing to inflation in the short-run, providing an impetus to aggregate demand, bank lending, and general economic activity by ensuring generous liquidity within the financial system. ≪/SYS≫ what traits suggest that the following text snippet (delimited by triple quotes) was being dovish? Text snippet: ```Input Sequence```. [/INST] The traits making it sound like dovish include"*

- Unified concept prompt (definition rephrased): *"[INST] ≪SYS≫ Pretend you are a financial analyst. Your task is to analyze the monetary policy stance of a text snippet from the Federal Open Market Committee (FOMC) document. Particularly, you should focus on what traits in the text snippet suggesting that this text is being hawkish or being dovish. hawkish is defined by more constraint in expanding diverse measures of money supply like M0, M1, M2, more worried about risks stemming from excess money supply contributing to inflation in the medium-term, implementing robust supervision on the monetary foundation served to shield purchasing power, elevate short-term borrowing costs and decrease valuations for assets such as stocks or housing. dovish is defined by* *more amenable to expanding the money supply at a vigorous pace through means such as quantitative easing, less worried about risks stemming from excess money supply contributing to inflation in the short-run, providing an impetus to aggregate demand, bank lending, and general economic activity by ensuring generous liquidity within the financial system. ≪/SYS≫ what traits suggest that the following text snippet (delimited by triple quotes) was being hawkish or being dovish? Text snippet: ```Input Sequence```. [/INST] The traits contribute to the monetary policy stance include"*

- Positive concept prompt (instruction rephrased): *"[INST] ≪SYS≫ Imagine you are a financial analyst tasked with analyzing the monetary policy stance of a sentence from the Federal Open Market Committee (FOMC) document. Specifically, your focus should be on identifying features of the sentence that contribute to its more hawkish tone. In more precise terms, sentences are considered hawkish if they are more restraint in growing various measures of money supply like M0, M1, M2, more concerned about risks of excess money supply expansion fueling inflation in the medium-term, maintaining strong controls on the monetary base to safeguard purchasing power, push up short-term borrowing costs and lowers asset valuations like stocks or housing. ≪/SYS≫ What aspects of the content in the following sentence (enclosed in triple quotes) might contribute to its more hawkish tone? Text snippet: ```Input Sequence```. [/INST] The traits making it sound like hawkish include"*

- Negative concept prompt (instruction rephrased): *"[INST] ≪SYS≫ Imagine you are a financial analyst tasked with analyzing the monetary policy stance of a sentence from the Federal Open Market Committee (FOMC) document. Specifically, your focus should be on identifying features of the sentence that contribute to its more dovish tone. In more precise terms, sentences are considered dovish if they are more willing to grow the money supply at a robust pace through measures like quantitative easing, less concerned about risks of too much money supply expansion fueling inflation in the short-run, giving a boost to aggregate demand, bank lending, and overall economic activity by ensuring ample liquidity in the financial system. ≪/SYS≫ What aspects of the content in the following sentence (enclosed in triple quotes) might contribute to its more dovish tone? Text snippet: ```Input Sequence```. [/INST] The traits making it sound like dovish include"*

- Unified concept prompt (instruction rephrased): *"[INST] ≪SYS≫ Imagine you are a financial analyst tasked with analyzing the monetary policy stance of a sentence from the Federal Open Market Committee (FOMC) document. Specifically, your focus should be on identifying features of the sentence that contribute to its more hawkish or dovish tone. In more precise terms, sentences are considered hawkish if they are more restraint in growing various measures of money supply like M0, M1, M2, more concerned about risks of excess money supply expansion fueling inflation in the medium-term, maintaining strong controls on the monetary base to safeguard purchasing power, push up short-term borrowing costs and lowers asset valuations like stocks or housing while they are considered dovish if they are more willing to grow the money supply at a robust pace through measures like quantitative easing, less concerned about risks of too much money supply expansion fueling inflation in the short-run, giving a boost to aggregate demand, bank lending, and overall economic activity by ensuring ample liquidity in the financial system. ≪/SYS≫ What aspects of the content in the following sentence (enclosed in triple quotes) might contribute to its more hawkish or dovish tone? Text snippet: ```Input Sequence```. [/INST] The traits contribute to the monetary policy stance include"*

***Study 2: Corporate innovation level:***

- Positive concept prompt (definition rephrased): *"[INST] ≪SYS≫ Pretend you are a financial analyst. Your task is to analyze the firm's innovation of a text snippet from the firm's earning conference call. Particularly, you should focus on what traits in the text snippet suggesting that this text is being innovative. innovative is defined by the generation or execution of fresh processes, products, or services by a company to drive forward, compete effectively, and distinguish themselves within the market. ≪/SYS≫ what traits suggest that the following text snippet (delimited by triple quotes) was being innovative? Text snippet: ```Input Sequence```. [/INST] The traits making it sound like innovative include"*

- Negative concept prompt (definition rephrased): *"[INST] ≪SYS≫ Pretend you are a financial analyst. Your task is to analyze the firm's innovation of a text snippet from the firm's earning conference call. Particularly, you should focus on what traits in the text snippet suggesting that this text is being conservative. conservative is defined by a hesitance to adopt daring innovations, resist digital disruption, or make significant* *departures from well-established products, processes, and strategies. ≪/SYS≫ what traits suggest that the following text snippet (delimited by triple quotes) was being conservative? Text snippet: ```Input Sequence```. [/INST] The traits making it sound like conservative include"*

- Unified concept prompt (definition rephrased): *"[INST] ≪SYS≫ Pretend you are a financial analyst. Your task is to analyze the firm's innovation of a text snippet from the firm's earning conference call. Particularly, you should focus on what traits in the text snippet suggesting that this text is being innovative or being conservative. innovative is defined the generation or execution of fresh processes, products, or services by a company to drive forward, compete effectively, and distinguish themselves within the market. conservative is defined by a hesitance to adopt daring innovations, resist digital disruption, or make significant departures from well-established products, processes, and strategies. ≪/SYS≫ what traits suggest that the following text snippet (delimited by triple quotes) was being innovative or being conservative? Text snippet: ```Input Sequence```. [/INST] The traits contribute to the firm's innovation include"*

- Positive concept prompt (instruction rephrased): *"[INST] ≪SYS≫ Imagine you are a financial analyst tasked with analyzing the firm's innovation of a text snippet from the firm's earning conference call. Specifically, your focus should be on identifying features of the text snippet that contribute to its more innovative tone. In more precise terms, text snippets are considered innovative if they are the creation or implementation of new processes, products or services by a firm to advance, compete and differentiate themselves in the market. ≪/SYS≫ what aspects of the content in the following text snippet (enclosed in triple quotes) might contribute to its more innovative tone? Text snippet: ```Input Sequence```. [/INST] The traits making it sound like innovative include"*

- Negative concept prompt (instruction rephrased): *"[INST] ≪SYS≫ Imagine you are a financial analyst tasked with analyzing the firm's innovation of a text snippet from the firm's earning conference call. Specifically, your focus should be on identifying features of the text snippet that contribute to its more conservative tone. In more precise terms, text snippets are considered conservative if they are a reluctance to embrace bold innovations, digital disruption or make substantial deviations from tried-and-true products, processes and strategies. ≪/SYS≫ what aspects of the content in the following* *text snippet (enclosed in triple quotes) might contribute to its more conservative tone? Text snippet: ```Input Sequence```. [/INST] The traits making it sound like conservative include"*

- Unified concept prompt (instruction rephrased): *"[INST] ≪SYS≫ Imagine you are a financial analyst tasked with analyzing the firm's innovation of a text snippet from the firm's earning conference call. Specifically, your focus should be on identifying features of the text snippet that contribute to its more innovative or conservative tone. In more precise terms, text snippets are considered innovative if they are the creation or implementation of new processes, products or services by a firm to advance, compete and differentiate themselves in the market while they are considered conservative if they are a reluctance to embrace bold innovations, digital disruption or make substantial deviations from tried-and-true products, processes and strategies. ≪/SYS≫ what aspects of the content in the following text snippet (enclosed in triple quotes) might contribute to its more innovative or conservative tone? Text snippet: ```Input Sequence```. [/INST] The traits contribute to the firm's innovation include"*

***Study 3: Information overload in online reviews:***

- Positive concept prompt (definition rephrased): *"[INST] ≪SYS≫ Pretend you are a linguist. Your task is to analyze the content complexity of a customer review about hotels. Particularly, you should focus on the characteristics regarding the review content that may make the review complex. More formally, a complex review is characterized by extensive and intricate assessments that cover a broad spectrum of elements, incorporate varied subjects, and offer subtle perspectives on the hotel stay. ≪/SYS≫ What characteristics regarding the content of the following customer review about hotels (delimited by triple quotes), do you think, may contribute to a complex information complexity? Review text: ```Input Sequence```. [/INST] The characteristics making the review complex include"*

- Negative concept prompt (definition rephrased): *"[INST] ≪SYS≫ Pretend you are a linguist. Your task is to analyze the content complexity of a customer review about hotels. Particularly, you should focus on the characteristics regarding the review content that may make the review concise. More formally, a concise review is characterized by straightforward and surface-level evaluations that lack profundity, concentrate on* *specific subjects, and provide restricted analysis of the hotel stay. ≪/SYS≫ What characteristics regarding the content of the following customer review about hotels (delimited by triple quotes), do you think, may contribute to a concise information complexity? Review text: ```Input Sequence```. [/INST] The characteristics making the review concise include"*

- Unified concept prompt (definition rephrased): *"[INST] ≪SYS≫ Pretend you are a linguist. Your task is to analyze the content complexity of a customer review about hotels. Particularly, you should focus on the characteristics regarding the review content that may make the review either complex or concise. More formally, a complex review is characterized by extensive and intricate assessments that cover a broad spectrum of elements, incorporate varied subjects, and offer subtle perspectives on the hotel stay. A concise review is characterized by straightforward and surface-level evaluations that lack profundity, concentrate on specific subjects, and provide restricted analysis of the hotel stay. ≪/SYS≫ What characteristics regarding the content of the following customer review about hotels (delimited by triple quotes), do you think, may contribute to the content complexity of this review, making it either complex or concise? Review text: ```Input Sequence```. [/INST] The characteristics contributing to its content complexity include"*

- Positive concept prompt (instruction rephrased): *"[INST] ≪SYS≫ Imagine you are a linguist tasked with analyzing the content complexity of a hotel customer review. Specifically, your focus should be on identifying features of the review that contribute to its complexity. In more precise terms, reviews are considered complex if they are comprehensive and detailed reviews that encompass a wide range of aspects, incorporate diverse topics, and provide nuanced insights into the hotel experience. ≪/SYS≫ What aspects of the content in the following customer hotel review (enclosed in triple quotes) might contribute to its complexity? Customer review: ```Input Sequence```. [/INST] The characteristics making the review complex include"*

- Negative concept prompt (instruction rephrased): *"[INST] ≪SYS≫ Imagine you are a linguist tasked with analyzing the content complexity of a hotel customer review. Specifically, your focus should be on identifying features of the review that contribute to its conciseness. In more precise terms, reviews are considered concise if they are* *simplistic and superficial reviews that lack depth, focus on narrow topics, and offer limited analysis of the hotel experience. ≪/SYS≫ What aspects of the content in the following customer hotel review (enclosed in triple quotes) might contribute to its conciseness? Customer review: ```Input Sequence```. [/INST] The characteristics making the review concise include"*

- Unified concept prompt (instruction rephrased): *"[INST] ≪SYS≫ Imagine you are a linguist tasked with analyzing the content complexity of a hotel customer review. Specifically, your focus should be on identifying features of the review that contribute to its complexity or conciseness. In more precise terms, reviews are considered complex if they are comprehensive and detailed reviews that encompass a wide range of aspects, incorporate diverse topics, and provide nuanced insights into the hotel experience, while they are considered concise if they are simplistic and superficial reviews that lack depth, focus on narrow topics, and offer limited analysis of the hotel experience. ≪/SYS≫ What aspects of the content in the following customer hotel review (enclosed in triple quotes) might contribute to its complexity or conciseness? Review text: ```Input Sequence```. [/INST] The characteristics contributing to its content complexity include"*

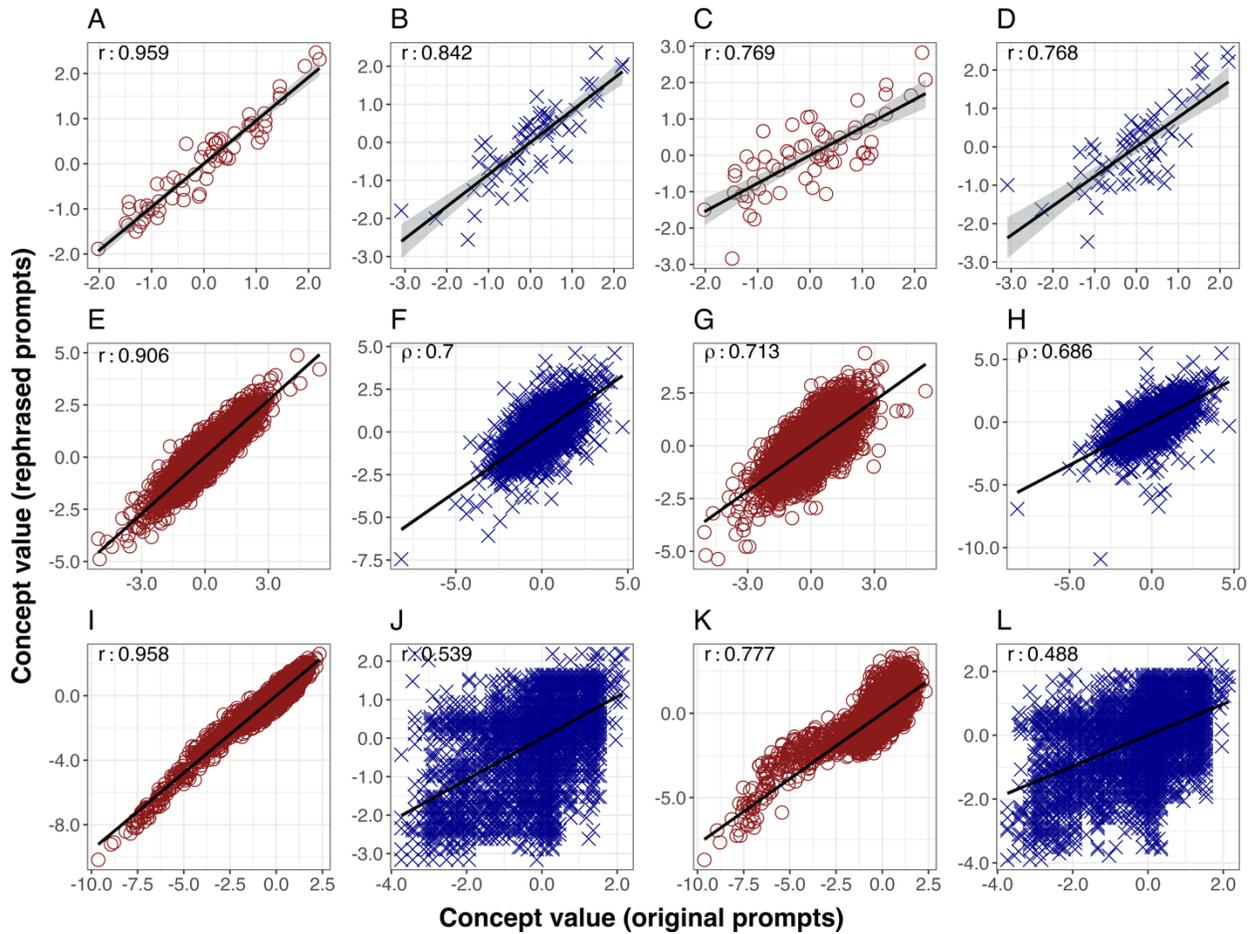

**Fig. S1. Correlation plots for sensitivity test.** We compare two sets of concept values (standardized) obtained using original (horizontal axis) and rephrased prompts (vertical axis). Each row corresponds to a concept being measured: the Fed's monetary policy stance (row 1); corporate innovation (row 2); and information overload in online reviews (row 3). Each column relates to a measurement method, with the 1st and 3rd columns using our LLM-Measure (**A**, **E**, **I**, **C**, **G**, **K** marked with red circles) and the 2nd and 4th columns using LLM-prompting (**B**, **F**, **J**, **D**, **H**, **L** marked with blue crosses). The first two columns, from left to right, consider rephrasing definitions, and the last two columns consider rephrasing instructions. We report Pearson correlation coefficient in the upper left corner of each plot, with all values being statistically significant ($P < 10^{-12}$).

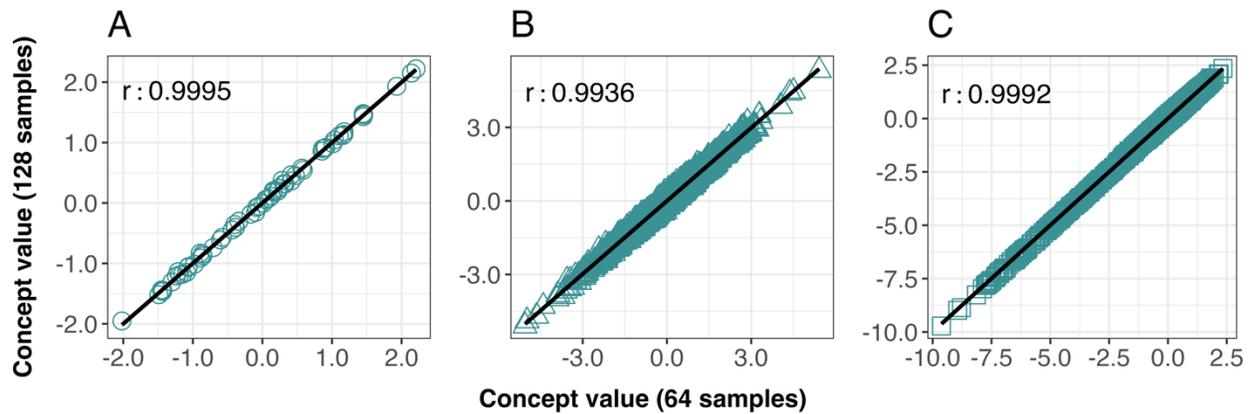

**Fig. S2. Correlation plots of concept values obtained with different probing sets.** The three charts compare the concept values obtained using different numbers of probing samples (64 vs. 128) in our three replication studies: (**A**) measuring the Fed's monetary policy stance, (**B**) measuring corporate innovation, and (**C**) measuring information overload in online reviews. We report Pearson correlation coefficient *r* in the upper left corner of each plot, with all values being statistically significant ($P < 10^{-15}$). The strong correlations offer evidence that 64 samples are sufficient to achieve stable measures.

**Table S1. Results of regressing U.S. Treasury yield on the Fed's monetary policy stance.** Note: *t-statistics* are reported below the estimates. Statistical significance at the 1%, 5%, and 10% levels are indicated by ***, **, and *, respectively.

|  | OLS model 1 ||| OLS model 2 ||| OLS model 3 |||
| --- | --- | --- | --- | --- | --- | --- | --- | --- | --- |
|  | 3-month Treasury yield ||| 1-year Treasury yield ||| 10-year Treasury yield |||
| Intercept | 0.712*** (6.397) | 0.712*** (6.064) | 0.712*** (6.549) | 0.904*** (7.900) | 0.904*** (7.071) | 0.904*** (8.395) | 2.054*** (24.044) | 2.054*** (23.055) | 2.054*** (26.677) |
| *PolicyStance* (RoBERTa) | 0.291** (2.613) |  |  | 0.452*** (3.949) |  |  | 0.252*** (2.949) |  |  |
| *PolicyStance* (LLM-prompting) |  | -0.030 (-0.255) |  |  | 0.096 (0.751) |  |  | 0.159* (1.780) |  |
| *PolicyStance* (LLM-Measure) |  |  | 0.344*** (3.171) |  |  | 0.542*** (5.036) |  |  | 0.382*** (4.957) |
| $R^2$ | 0.102 | 0.001 | 0.144 | 0.206 | 0.009 | 0.297 | 0.127 | 0.05 | 0.291 |
| N | 62 | 62 | 62 | 62 | 62 | 62 | 62 | 62 | 62 |

**Table S2. Results of regressing R&D spending on corporate innovation.** Note: *t-statistics* are reported below the estimates. Statistical significance at the 1%, 5%, and 10% levels are indicated by ***, **, and *, respectively.

|  | OLS model | | |
|---|---|---|---|
|  | R&D expenses divided by total assets | | |
| Intercept | 0.378*** (34.966) | 0.365*** (33.979) | 0.378*** (35.263) |
| *Innovation* (Dictionary) | 0.010*** (3.193) | | |
| *Innovation* (LLM-prompting) | | 0.035*** (11.408) | |
| *Innovation* (LLM-Measure) | | | 0.031*** (10.152) |
| *Size* | -0.0395*** (-27.226) | -0.038*** (-26.124) | -0.039*** (-27.453) |
| $R^2$ | 0.125 | 0.145 | 0.140 |
| N | 5,237 | 5,237 | 5,237 |

**Table S3. Results of regressing helpfulness on information overload of online reviews.** Note: *t-statistics* are reported below the estimates. Statistical significance at the 1%, 5%, and 10% levels are indicated by ***, **, and *, respectively.

|  | OLS model 1 | | | OLS model 2 | | |
|---|---|---|---|---|---|---|
|  | $log\ (Helpfulness\ +\ 1)$ | | | $Helpfulness$ | | |
| Intercept | -0.376*** (-7.727) | -0.244*** (-4.633) | -0.454*** (-9.078) | -1.586*** (-8.040) | -1.213*** (-5.684) | -1.941*** (-9.572) |
| $InformationOverload$ (LDA) | -0.018** (-2.553) | | | -0.054* (-1.894) | | |
| $InformationOverload$ (LLM-prompting) | | 0.024*** (3.095) | | | 0.066** (2.078) | |
| $InformationOverload$ (LLM-Measure) | | | -0.044*** (-5.696) | | | -0.175*** (-5.567) |
| $Sentiment$ | -0.113*** (-7.562) | -0.112*** (-7.475) | -0.132*** (-8.638) | -0.332*** (-5.454) | -0.329*** (-5.401) | -0.404*** (-6.520) |
| $Words$ | 0.207*** (21.881) | 0.179*** (17.123) | 0.226*** (22.559) | 0.649*** (16.949) | 0.570*** (13.473) | 0.735*** (18.095) |
| $R^2$ | 0.063 | 0.063 | 0.065 | 0.038 | 0.038 | 0.041 |
| $N$ | 10,000 | 10,000 | 10,000 | 10,000 | 10,000 | 10,000 |

**Table S4. Summary statistics for variables used in the regression analysis for study 1.**

| Variable | Description | Mean | Median | Min-max | St. dev. |
|---|---|---|---|---|---|
| *Yield* (3-month) | Daily yield of treasury with 3-month maturity | 0.71 | 0.11 | 0.01–3.31 | 0.91 |
| *Yield* (1-year) | Daily yield of treasury with 1-year maturity | 0.90 | 0.28 | 0.05–4.08 | 0.99 |
| *Yield* (10-year) | Daily yield of treasury with 10-year maturity | 2.05 | 2.02 | 0.58–3.51 | 0.71 |
| *PolicyStance* (RoBERTa) | Hawkishness of policy stance from *Shah et al.* (*13*) | -0.04 | -0.08 | -0.49–0.53 | 0.23 |
| *PolicyStance* (LLM-prompting) | Hawkishness of policy stance by LLM-prompting | 54.23 | 54.31 | 41.97–62.98 | 3.94 |
| *PolicyStance* (LLM-Measure) | Hawkishness of policy stance by LLM-Measure | -0.04 | -0.02 | -0.57–0.53 | 0.26 |

**Table S5. Summary statistics for variables used in the regression analysis for study 2.**

| Variable | Description | Mean | Median | Min-max | St. dev. |
|---|---|---|---|---|---|
| *R&Dspending* | R&D expenditures normalized by total assets | 0.10 | 0.03 | 0–10.83 | 0.24 |
| *Size* | Natural logarithm of total assets | 7.15 | 7.24 | 0.66–13.30 | 2.15 |
| *Innovation* (Dictionary) | Innovation measure from Li *et al.* (*14*) | 6.15 | 5.35 | 0.44–27.39 | 3.43 |
| *Innovation* (LLM-prompting) | Innovation measure by LLM-prompting | 66.56 | 66.48 | 31.00–87.00 | 4.35 |
| *Innovation* (LLM-Measure) | Innovation measure by LLM-Measure | 0.10 | 0.09 | -2.94–3.36 | 0.60 |

**Table S6. Summary statistics for variables used in the regression analysis for study 3.**

| Variable | Description | Mean | Median | Min-max | St. dev. |
|---|---|---|---|---|---|
| $Rating$ | Overall rating on a five-star scale | 3.97 | 4.00 | 1–5 | 1.15 |
| $Helpfulness$ | Number of votes a review receives | 1.26 | 0.00 | 0–74 | 2.66 |
| $Words$ | Number of words in a review | 154.32 | 120.00 | 7–2383 | 128.91 |
| $Sentiment$ | Binary score (0 for ratings of three stars or fewer and 1 for ratings of four or five stars) | 0.74 | 1.00 | 0–1 | 0.44 |
| $log(Words)$ | Natural logarithm of the number of words in a review | 4.76 | 4.79 | 1.94–7.78 | 0.76 |
| $log(Helpfulness + 1)$ | Natural logarithm of one plus the number of votes a review receives | 0.52 | 0.00 | 0–4.32 | 0.66 |
| $InformationOverload$ (LDA) | Information overload as topic entropy by LDA | 1.26 | 1.29 | 0.05–2.45 | 0.36 |
| $InformationOverload$ (LLM-prompting) | Information overload by LLM-prompting | 62.28 | 65.00 | 0–98 | 16.67 |
| $InformationOverload$ (LLM-Measure) | Information overload by LLM-Measure | 107.18 | 107.35 | 94.40–110.28 | 1.33 |

**Table S7. Selected reviews with the highest and lowest information overload values (top and bottom three).** Projection values are shown in parentheses above each sample.

(−9.628)
I highly recommend this hotel to anyone visiting NY! It was amazing!

(−8.939)
Spent on night on a business trip recently. Nvery nice hotel.

(−8.781)
Great experience\ni suggest this hotel to everyone

(+2.341)
I come to NYC fairly often, mostly for business and we enjoyed our week stay here. I stayed at the Marriott East Side many times in the past, but will now stay at this hotel because it is ideally located, its large rooms, friendly staff and price. The location is perfect (6th Ave. and W. 38th St.) since the hotel is less than a block from the subway and about a five minute walk to Times Square. \nSome reviewers said they had problems with the elevators, but we never had any problems with the elevators or crowds with the breakfast buffet (they do leave you a voicemail on the weekends to get in early to avoid crowds, but didn't have any problems at 9 am on a Saturday). Rooms were huge and especially clean by NYC standards and having the kitchen for leftovers saved us a mint. The manager, Ryan, and the entire staff were outstanding. Ask for a room on a high floor - 32, 33, or 34 and on the SE side so you have a view of the Empire State Building - there's only about 6 or 7 rooms per floor, so it may be worth calling in advance for special requests.\nSome tips for people looking for restaurants and a music hot spot that are nearby or only a short subway ride: New York Burger Co. (As good as Shake Shack and no long lines) on 678 6th Ave. between 21st.and 22nd St.), Levain Bakery - I know it's a bakery, but it has the best cookies you'll ever find - 167 W 74TH St (Amsterdam cross street) Central Park Boathouse Restaurant - Great spot for the weekends when it's really nice outside, Carmine's (close to Times Square on W. 44th between Broadway and 8th Ave., Cafe Wha? - Don't go here for the food, go here for the music - best you'll find in NYC, located in Greenwich Village.

(+2.177)
Pretentious. They are more concerned about looking cool than providing good service or caring for guests. Overall we were disappointed. For $600+ / N we expected better; we got less than mediocre. Not the best standard to strive for. Focus on the avg or negative reviews on Tripadvisor as they seem more genuine (i'm suspicious about the glowing reviews as certain hotels in NY have a rep for \"padding\"). Can do much better at this price range. Concierge was useless - did not have any local knowledge and referred to his Zagats for his recommendations (we could have done that!!).

(+2.160)
I used Tripadvisor to review various properties at the Philadelphia airport before booking and the Fairfield Inn was highly rated so we booked. We were astonished to find that the reviews in no way reflect reality. \nI want to say first of all that the staff were professional, competent and extremely helpful, so this review in no way reflects on them. While the lobby and breakfast room were clean in appearance, the room was a completely different matter. It was shabby and filthy. The sheets and towels were clean but the furniture was sticky, the shower curtain had mold around the edges, and the chair in the room had multiple stains on the seat. The curtains and dust ruffles around the beds were very wrinkled and appeared to have been crumpled in a ball before being used. After noticing these things, I began looking more closely at other aspects of the hotel and noticed that the elevators were quite old and had not been cleaned well either. The hallway carpets appeared dirty and were rippled from stretching - they are obviously old as well and not kept up.\nThis is the second worst hotel I've stayed in and I've stayed in a lot of hotels all over the world. I would never recommend this hotel to anyone or stay here or in another Fairfield Inn again. When you can't even sit on the chair in the room because it is so badly stained or touch the tops of the furniture because your fingers will stick to the top, it is pretty bad.

## References

1. M. Gentzkow, B. Kelly, M. Taddy, Text as data. *Journal of Economic Literature* **57**, 535–574 (2019).

2. M. E. Roberts, B. M. Stewart, E. M. Airoldi, A model of text for experimentation in the social sciences. *Journal of the American Statistical Association* **111**, 988–1003 (2016).

3. J. Wilkerson, A. Casas, Large-scale computerized text analysis in political science: Opportunities and challenges. *Annual Review of Political Science* **20**, 529–544 (2017).

4. S. Rathje, D.-M. Mirea, I. Sucholutsky, R. Marjieh, C. E. Robertson, J. J. Van Bavel, GPT is an effective tool for multilingual psychological text analysis. *Proceedings of the National Academy of Sciences* **121**, e2308950121 (2024).

5. G. Le Mens, B. Kovács, M. T. Hannan, G. Pros, Uncovering the semantics of concepts using GPT-4. *Proceedings of the National Academy of Sciences* **120**, e2309350120 (2023).

6. F. Gilardi, M. Alizadeh, M. Kubli, ChatGPT outperforms crowd workers for text-annotation tasks. *Proceedings of the National Academy of Sciences* **120**, e2305016120 (2023).

7. M. Gentzkow, J. M. Shapiro, M. Taddy, Measuring group differences in high-dimensional choices: method and application to congressional speech. *Econometrica* **87**, 1307–1340 (2019).

8. C. A. Bail, Can Generative AI improve social science? *Proceedings of the National Academy of Sciences* **121**, e2314021121 (2024).

9. M. Sclar, Y. Choi, Y. Tsvetkov, A. Suhr, Quantifying Language Models' Sensitivity to Spurious Features in Prompt Design or: How I learned to start worrying about prompt formatting. *arXiv:2310.11324* (2023).

10. A. Zou, L. Phan, S. Chen, J. Campbell, P. Guo, R. Ren, A. Pan, X. Yin, M. Mazeika, A.-K. Dombrowski, Representation engineering: A top-down approach to ai transparency. *arXiv:2310.01405* (2023).

11. H. Touvron, L. Martin, K. Stone, P. Albert, A. Almahairi, Y. Babaei, N. Bashlykov, S. Batra, P. Bhargava, S. Bhosale, Llama 2: Open foundation and fine-tuned chat models. *arXiv:2307.09288* (2023).

12. T. Brown, B. Mann, N. Ryder, M. Subbiah, J. D. Kaplan, P. Dhariwal, A. Neelakantan, P. Shyam, G. Sastry, A. Askell, S. Agarwal, A. Herbert-Voss, G. Krueger, T. Henighan, R. Child, A. Ramesh, D. Ziegler, J. Wu, C. Winter, C. Hesse, M. Chen, E. Sigler, M. Litwin, S. Gray, B. Chess, J. Clark, C. Berner, S. McCandlish, A. Radford, I. Sutskever, D. Amodei, "Language Models are Few-Shot Learners" in *Proceedings of the 34th International*

*Conference on Neural Information Processing Systems* (Curran Associates Inc., 2020), pp. 1877–1901.

13. A. Shah, S. Paturi, S. Chava, "Trillion Dollar Words: A New Financial Dataset, Task & Market Analysis" in *Proceedings of the 61st Annual Meeting of the Association for Computational Linguistics* (Association for Computational Linguistics, 2023), pp. 6664–6679.

14. K. Li, F. Mai, R. Shen, X. Yan, Measuring Corporate Culture Using Machine Learning. *The Review of Financial Studies* **34**, 3265–3315 (2021).

15. A. Ghose, P. G. Ipeirotis, B. Li, Modeling consumer footprints on search engines: An interplay with social media. *Management Science* **65**, 1363–1385 (2019).

16. B. S. Bernanke, A. S. Blinder, The Federal Funds Rate and the Channels of Monetary Transmission. *American Economic Review* **82**, 901–921 (1992).

17. S. Hansen, M. McMahon, A. Prat, Transparency and deliberation within the FOMC: A computational linguistics approach. *The Quarterly Journal of Economics* **133**, 801–870 (2018).

18. Y. Liu, M. Ott, N. Goyal, J. Du, M. Joshi, D. Chen, O. Levy, M. Lewis, L. Zettlemoyer, V. Stoyanov, Roberta: A robustly optimized bert pretraining approach. *arXiv:1907.11692* (2019).

19. J. A. Schumpeter, *Capitalism, Socialism and Democracy* (routledge, 2013).

20. P. Aghion, P. W. Howitt, *The Economics of Growth* (MIT press, 2008).

21. J. Jacoby, Perspectives on information overload. *Journal of consumer research* **10**, 432–435 (1984).

22. D. M. Blei, A. Y. Ng, M. I. Jordan, Latent dirichlet allocation. *Journal of machine Learning research* **3**, 993–1022 (2003).

23. B. A. Nosek, G. Alter, G. C. Banks, D. Borsboom, S. D. Bowman, S. J. Breckler, S. Buck, C. D. Chambers, G. Chin, G. Christensen, Promoting an open research culture. *Science* **348**, 1422–1425 (2015).

24. V. Stodden, M. McNutt, D. H. Bailey, E. Deelman, Y. Gil, B. Hanson, M. A. Heroux, J. P. Ioannidis, M. Taufer, Enhancing reproducibility for computational methods. *Science* **354**, 1240–1241 (2016).

25. S. Santurkar, E. Durmus, F. Ladhak, C. Lee, P. Liang, T. Hashimoto, "Whose opinions do language models reflect?" in *Proceedings of the 40th International Conference on Machine Learning* (PMLR, 2023), pp. 29971–30004.

26. I. Grossmann, M. Feinberg, D. C. Parker, N. A. Christakis, P. E. Tetlock, W. A. Cunningham, AI and the transformation of social science research. *Science* **380**, 1108–1109 (2023).